\documentclass{elsarticle}

\usepackage{lineno,hyperref}
\modulolinenumbers[5]

\journal{Journal of XXXXX}
\usepackage{amssymb,amsmath,amsfonts}
\usepackage{epsfig}
\usepackage{mathptmx}
\usepackage{graphicx}
\usepackage{color}
\usepackage{caption}
\usepackage{wrapfig}
\usepackage{graphics}
\usepackage{enumerate}
\usepackage{amsxtra}
\usepackage{latexsym}
\usepackage{multirow}
\usepackage{slashbox}
\usepackage{subfigure}
\usepackage{epstopdf}
\usepackage{pdfpages}
\usepackage{algorithm}
\usepackage{algorithmic}
\usepackage{url}
\newcommand{\thmlist}{
\begin{list}{Step 1}
{\setlength{\leftmargin}{0.6 in}\setlength{\labelwidth} {0.5 in}}}
\newcommand{\alglist}{
\begin{list}{Step 1}
{\setlength{\leftmargin}{1.1 in} \setlength{\labelwidth}{1.0 in}}}
 \newcommand{\proof} {\noindent {\bf Proof.} \quad}
 \newcommand{\eproof} {$\quad \square$}
 
 \newtheorem{theorem}{Theorem}[section]
 
 \newtheorem{lemma}{Lemma}[section]

\newcommand{\subtitle}[1]{\color{blue}}

\mark{\indent Luo, Lv and Sun, Visual-inertial Navigation Method for UAVs}

\begin{document}

\begin{frontmatter}

\title{A Visual-inertial  Navigation Method for High-Speed Unmanned Aerial Vehicles
}
\author{
 Xin-long Luo \textsuperscript{a} \thanks{$^\ast$Corresponding
author: Xin-long Luo. Email: luoxinlong@bupt.edu.cn.},
Jia-hui Lv \textsuperscript{a} and  Geng Sun \textsuperscript{b}
\\
\textsuperscript{a} School of Information and Communication Engineering, \\
Beijing University of Posts and Telecommunications, P. O. Box 101, \\
Xitucheng Road  No. 10, Haidian District, 100876, Beijing China \\
luoxinlong@bupt.edu.cn, jhlv@bupt.edu.cn \\
\textsuperscript{b}Institute of Mathematics,
Academy of Mathematics and Systems Science, \\
Chinese Academy of Sciences, 100190, Beijing China \\
sung@amss.ac.cn
}

% \maketitle

\begin{abstract}
  This paper investigates the localization problem of high-speed high-altitude
  unmanned aerial vehicle (UAV) with a monocular camera and inertial navigation system.
  And it proposes a navigation method utilizing the complementarity
  of vision and inertial devices to overcome the singularity which arises from
  the horizontal flight of UAV. Furthermore, it modifies the mathematical model
  of localization problem via separating linear parts from nonlinear parts and
  replaces a nonlinear least-squares problem with a linearly equality-constrained
  optimization problem. In order to avoid the ill-condition property near
  the optimal point of sequential unconstrained minimization techniques
  (penalty methods), it constructs a semi-implicit continuous method with
  a trust-region technique based on a differential-algebraic dynamical system
  to solve the linearly equality-constrained optimization problem. It also analyzes
  the global convergence property of the semi-implicit continuous method in an
  infinity integrated interval other than the traditional convergence analysis of
  numerical methods for ordinary differential equations in a finite integrated
  interval. Finally, the promising numerical results are also presented.
\end{abstract}



\end{frontmatter}

% \linenumbers
% main text

\section{Introduction}

\vskip 2mm

Localization is essential for autonomous navigation of unmanned aerial vehicles.
In terms of aircraft navigation, the aircraft usually uses an inertial integrated
navigation method to guide flight due to the unsatisfactory effect of pure inertial
navigation system (INS) (see \cite{W2007}). The visual-based navigation method
has received widespread attention in recent years for its great performance in
this field. Therefore, we utilize visual odometer which is complementary to inertial
measurement to assist INS.

\vskip 2mm

Visual-based methods are often applied in low speed and low height situations,
resulting from such drawback as motion blur. Being different from others, our work
tries to solve the localization problem of high-speed unmanned aerial vehicles
(UAVs) in high altitude. In order to overcome the motion blur and scale ambiguity
that arise in this challenging practical problem, a novel visual based method
combining the inertial navigation system
is designed.

\vskip 2mm

In the case of challenging camera dynamics, the imaging of landmarks inevitably
appears blurred. According to the principle of pinhole imaging, we consider
an additional error to the angle of view for the landmarks imaging and the optical
center of camera lens, which is determined by the property of the camera. Note that
the angular error has a great impact on the position estimation of the aircraft due
to the extremely high flight altitude. For the scale ambiguity, rather than assume
the homography, we use the altitude difference of the aircraft in a short
time interval measured by altimeter to determine the height of aerial vehicle.

\vskip 2mm

For a real engineering problem, a UAV flight trajectory is usually relatively
simple. In particular, when a UAV flies on a horizontal plane, the flight altitude
difference with reading error within a short time interval will be intensely
small. Generally, that scenario leads to the singularity even using an altimeter
to help determine scale. The singularity results in rapid accumulation of errors.
In order to overcome its singularity, we add an inertial distance between two
sequential frames to assist visual localization.

\vskip 2mm

Furthermore, we modify the mathematical model of visual-inertial localization
problem via separating linear parts from nonlinear
parts and replace a nonlinear least-squares problem with a linearly
equality-constrained optimization problem. In order to avoid the ill-condition
property near the optimal point of sequential unconstrained minimization
techniques (penalty methods \cite{FM1990, SY2006}), we construct a semi-implicit
continuous method with a trust-region technique based on a differential-algebraic
dynamical system to solve the linearly equality-constrained optimization problem.
We also analyze the global convergence property of the proposed semi-implicit
continuous method in an infinity integrated interval other than the traditional
convergence analysis of numerical methods for ordinary differential equations
in a finite integrated interval.

\vskip 2mm

Finally, in order to validate the effectiveness of our proposed method, we adopt
real parameters provided by China Aerospace Science and Industry Corporation to
mimic the real flight environment and compare it with the pure inertial navigation
method \cite{YG2015}. The simulation results show that our proposed method has
better performance, and it meets the required accuracy in a long-term flight.

\vskip 2mm

The rest of the paper is organized as follows. Firstly, we present
the related work in section 2. Then, applied environment and sensors fusion
architecture of our visual-inertial navigation method are described in section 3.
In section 4, we modify the mathematical model and give a semi-implicit continuous
method with a trust-region technique to solve that optimization problem.
The simulation results of our method in comparison to the pure inertial navigation
method are presented in section 5. Finally, we give some discussions in section 6.

\vskip 2mm

\section{Related Work}

\vskip 2mm

In recent years, numerous methods have been applied to improve the precision
of navigation. Among of navigation systems, the strap-down inertial navigation
system (SINS) has great performance on pose estimation for its advantages of
complete autonomy, strong anti-interference ability and high short-term precision
\cite{TW2004, GWA2011, YLG2015}. However, the inertial measurement unit (IMU)
which is the main component of SINS has an unavoidable cumulate error caused
by sensor drifts \cite{AGKK2013, FLD2016}. Therefore, many aerial vehicles utilize
the position signal from GPS which fuses data generated by IMU to implement high
navigation accuracy.

\vskip 2mm

On the other hand, in many scenarios, GPS is difficult to play its role.
In civil applications, GPS becomes inaccurate due to multipath effect as close to
buildings and obstacles. In military field, such as ballistic missile vehicles,
the position applications generally do not rely on GPS by reason of jamming and
spoof \cite{EBM2018}. In GPS denied environments, vision-based approach is an
available and effective method, and visual-inertial odometer is ubiquitously
applied on robots and unmanned aerial vehicles (UAVs) for its great performance
of pose estimation and the complementarity between cameras and IMU
\cite{CD2008, CD2009, EBM2018, ESC2014, OBLS2015, QPTH2019, STF2017, ZS2015}.

\vskip 2mm

For the high altitude problem, we choose a monocular vision odometer to assist
INS rather than the binocular camera, since the binocular camera reduces to a
monocular camera when vehicles fly at high altitude as a result of the extremely
small baseline-to-depth ratio \cite{WU2013}. If there is no additional information,
scale ambiguity of a monocular camera can not be cleared up generally.
Anwar et al. design a new depth-independent Jacobian matrix by relating the depth
information with the area of region of interest \cite{ALDQG2019}. In \cite{CD2009},
Conte and Doherty consider the ground as flat and horizontal for aircraft flying at
a relatively high altitude. Caballero et al. also assume the local ground flat but
not level \cite{CMFO2009}. Both of them utilize planer homography to tackle the
vehicle motion. In this paper,
we give a method which does not require the local ground flat and does not
rely on planar homography.

\vskip 2mm

Zhang and Singh combine a high-accuracy INS and vision to estimate the position
of a full-scale aircraft flying at an altitude of about 300 meters. They partially
eliminate the effects of INS high-frequency noise through virtually rotating the
camera parallel to local ground by reparametrizing features with their depth
direction perpendicular to the ground \cite{ZS2015}. Unlike Zhang and Singh, our
method deal with the singular problem in the special case where the position
of two frames have no altitude difference. In addition, we separate the nonlinear
terms from the linear terms, and convert it to a linearly constrained optimization
problem, rather than directly adopt the Levenberg-Marquardt method to solve nonlinear
least-square problem. Furthermore, for that linearly equality-constrained optimization
subproblem, we give a semi-implicit continuous method with a
trust-region technique to solve it.

\vskip 2mm

\section{Sensors Fusion Architecture}

\vskip 2mm

\begin{figure}[htbp]
    \centering
    \includegraphics[width=4.5in,height =3in]{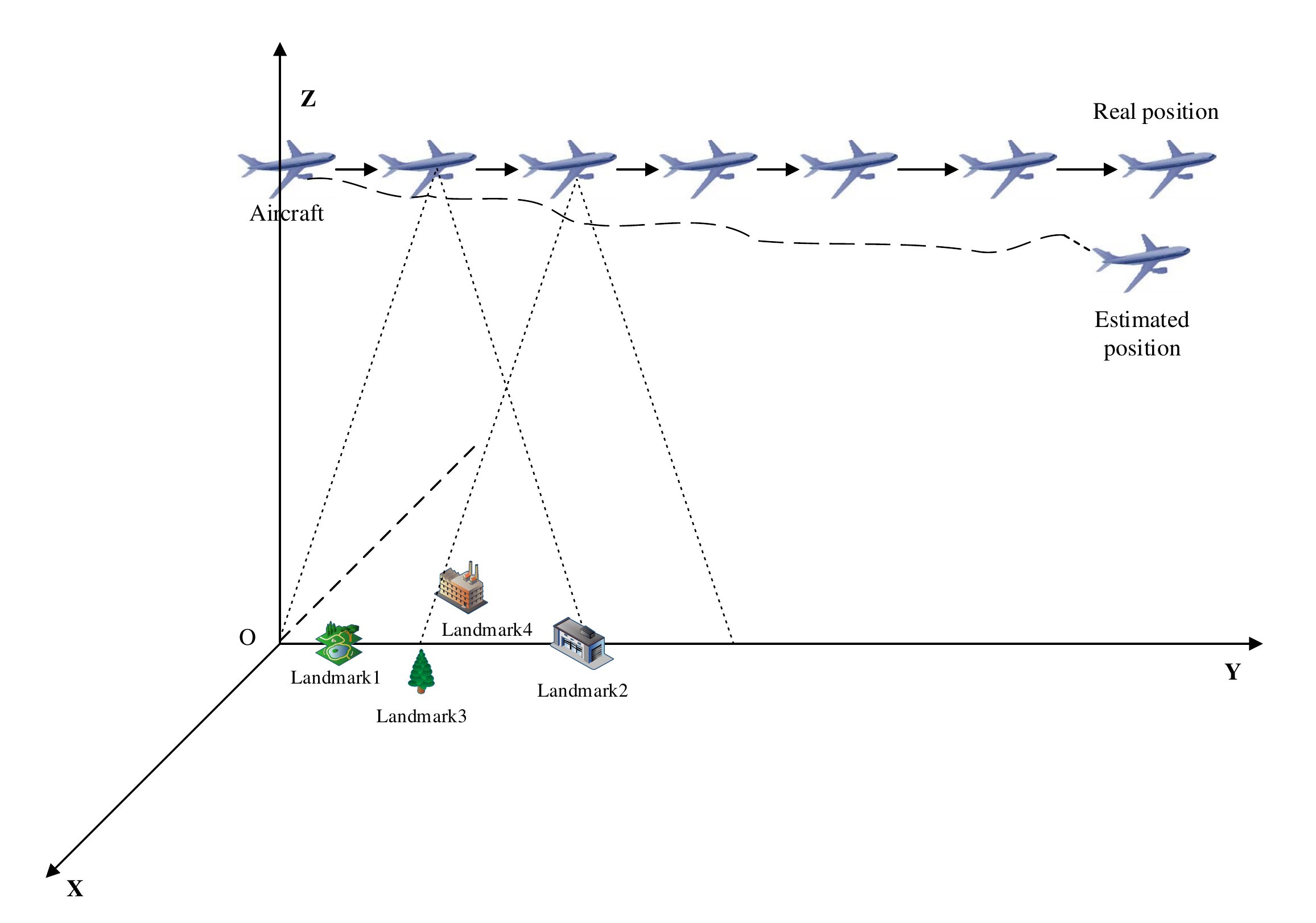}

    \caption{Navigation Simulation of Straight Line}
    \label{Fig:NSSL}
\end{figure}

In this paper, we focus our attention on the issue: solving the navigation problem
of high speed and high altitude aircraft under the horizontal flight scenario. The
navigation simulation is illustrated by Figure \ref{Fig:NSSL}.

\vskip 2mm

The visual-inertial odometry, which is composed by a monocular
vision system, an INS and an altimeter, is aimed to estimate the aircraft
position and guide the flight of aerial vehicles. We consider the camera fitting
a pinhole model briefly, and ignoring the lens distortion \cite{HZ2004}. The
Camera intrinsics parameters are given. As a convention, the 3D coordinate system
denotes the real world as shown in Figure \ref{Fig:NSSL}, and the symbol
$k, \; k \in Z^{+}$ denotes image frames. Besides the image coordinate system,
another coordinate system is a 2D coordinate system with its origin being
perpendicular to the optical center of camera lens, as shown in
Figure \ref{Fig:PM}. In the relatively difficult practical issue, velocity
of the aircraft is between 200 meters per second and 300 meters per second,
and the aerial vehicle flies at an altitude between 1000 meters and 1500 meters.
Since the aerial vehicle flies with an extremely fast speed, the motion blur
should be carefully considered. In order to reduce the influence of the blur,
we use a camera to assist localization and add an angular error to the angle of
view between camera and landmarks, about 0.2 degrees.

\vskip 2mm

\begin{figure}[http]
    \centering
    \includegraphics[width=4in,height =2in]{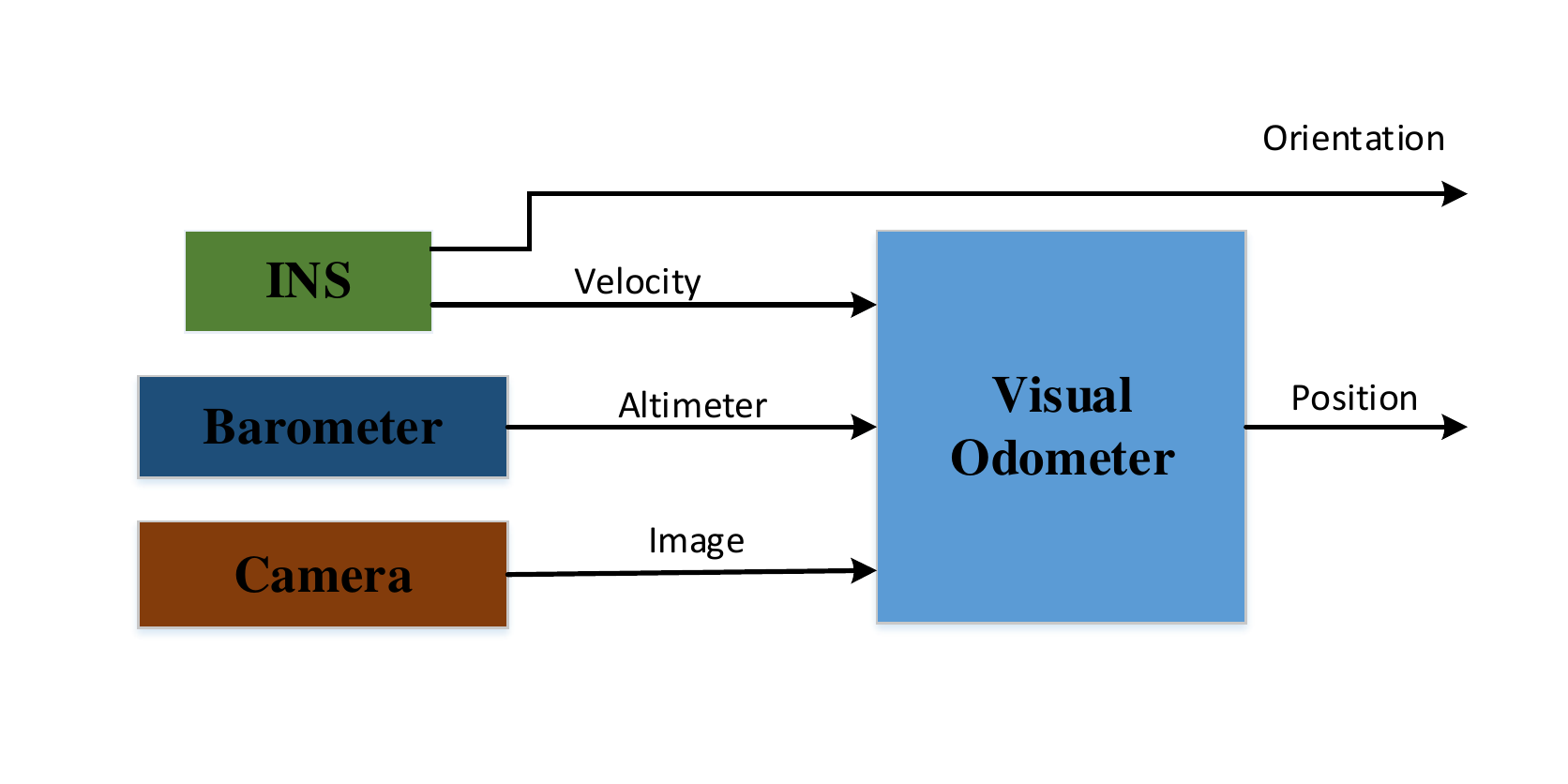}
    \vspace{-1cm}
    \caption{Sensor Fusion Architecture}
    \label{Fig:SFA}
\end{figure}

The sensor fusion architecture of the visual-inertial odometry is demonstrated
in Figure \ref{Fig:SFA}. The odometer takes the camera images, altimeter reading
from the altimeter and velocity from the INS. Combining those information, our
method can acquire the vehicle position with low drift in the horizonal flight.
When the aerial vehicle adjusts its orientation, the angle of rotation is obtained
by the IMU. Through acquiring the navigation information, reaching the destination
along the scheduled route can be achieved with required accuracy.

\vskip 2mm

In order to improve the navigation accuracy of ballistic missiles, we propose a method
which utilizes monocular camera to assist the inertial navigation system. A monocular
vision odometer typically has scale ambiguity. This scale ambiguity can be confirmed
by the barometer through measuring the flying altitude. On the other hand, a vision
odometer can suppress the cumulate error caused by IMU drift. Generally, taking full
advantage of the complementarity of visual odometer and inertial navigation, the
accuracy of navigation is improved.

\vskip 2mm

\section{Mathematical Model and Algorithm Descriptions}

\vskip 2mm

\subsection{Mathematical Model}

\begin{figure}[htbp]
    \centering
    \includegraphics[width=4in,height =3in]{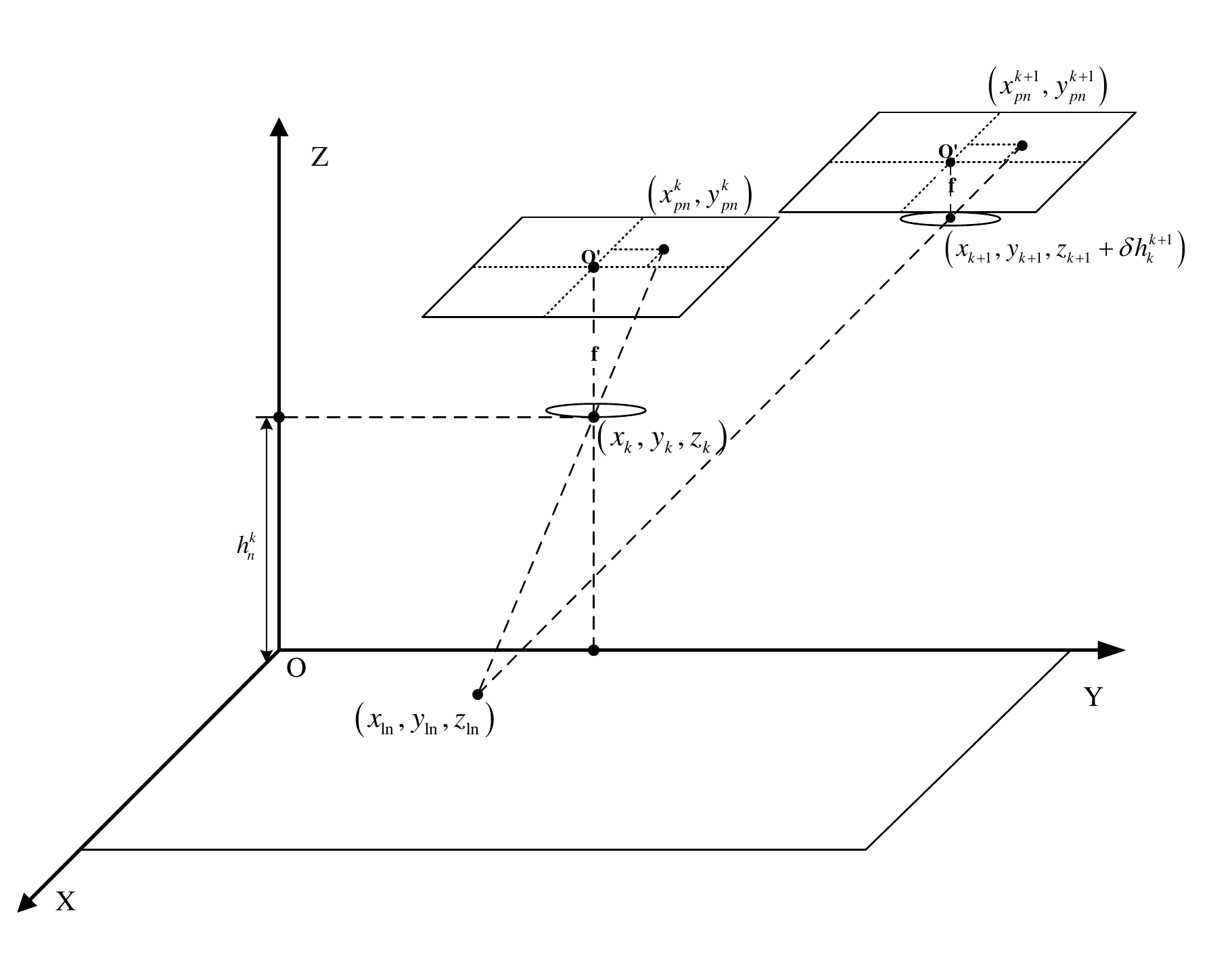}
    \caption{Mathematical model of visual-inertial odometer}
    \label{Fig:PM}
\end{figure}

This subsection is aimed to illustrate the mathematical model of localization
which is abstracted from practical problems. In that mathematical model, the
positions of camera and landmarks are in the world coordinate. The sequence
of frames is presented in parallel coordinate. Let the position of optical
center in the $k^{th}$ frame as $(x_{k},\, y_{k}, \, z_{k})$, and the $(k+1)^{th}$
frame as $(x_{k+1},\, y_{k+1}, \, z_{k+1}+\delta{h_{k}^{k+1}})$, where
$\delta{h_{k}^{k+1}}$ is the height difference between two frames, obtained
by an altimeter. The position of the $n$-th landmark, confirmed by ORB feature
\cite{EVKG2011}, is denoted as $(x_{ln},\, y_{ln}, \, z_{ln})$.
We denote the location of the corresponding pixel imaged by the $n^{th}$ landmark
in the $k^{th}$ frame as $(x_{pn}^{k},\, y_{pn}^{k})$. The vertical distance between
the $n^{th}$ landmark and the optical center of the camera of $k^{th}$ frame is
expressed as $h_{n}^{k}$. $f_c$ is the focal length of the camera. The precedent
notations are shown in Figure \ref{Fig:PM}. Figure \ref{Fig:PM} presents the
mathematic model of our visual odometer method appropriately.

\vskip 2mm

This subsection shows the mathematical derivation of our proposed visual-inertial
odometer method. In subsection \ref{SubSecAlg}, the complete algorithm is presented.
From the model in Figure \ref{Fig:PM}, we find the following relationships between
the $n^{th}$ landmark and the corresponding projection in two frames:
\begin{equation} \label{RLT:LIN}
\left\{
    \begin{split}
        \begin{array}{lr}
            \frac{x_{k} - x_{ln}}{h_{n}^{k}} &= \frac{x_{pn}^{k}}{f_c},\\
            \frac{y_{k} - y_{ln}}{h_{n}^{k}} &= \frac{y_{pn}^{k}}{f_c},\\
            \frac{x_{k+1} - x_{ln}}{h_{n}^{k} + \delta{h_{k}^{k+1}}}
            &= \frac{x_{pn}^{k+1}}{f_c},\\
            \frac{y_{k+1} - y_{ln}}{h_{n}^{k} + \delta{h_{k}^{k+1}}}
            &= \frac{y_{pn}^{k+1}}{f_c}.
        \end{array}
    \end{split}
\right.
\end{equation}
The above relationship (\ref{RLT:LIN})  is reformulated by
\begin{equation} \label{EQN:LIN}
    \left\{
    \begin{array}{lllcr}
        x_{ln} + \frac{x_{pn}^{k}}{f_c}h_{n}^{k} & = x_{k},\\
        y_{ln} + \frac{y_{pn}^{k}}{f_c}h_{n}^{k} & = y_{k},\\
        x_{k+1} - x_{ln} - \frac{x_{pn}^{k+1}}{f_c}h_{n}^{k}
        & = \frac{\delta{h_{k}^{k+1}}}{f_c}x_{pn}^{k+1},\\
        y_{k+1} - y_{ln} - \frac{y_{pn}^{k+1}}{f_c}h_{n}^{k}
        & = \frac{\delta{h_{k}^{k+1}}}{f_c}y_{pn}^{k+1}.
    \end{array}
    \right.
\end{equation}

\indent In formula (\ref{EQN:LIN}), the position of pixel in the camera
coordinate and the position of the $k^{th}$ frame are known, and the rest are
unknown. Obviously, this is an underdetermined system and we can not determine
the position of the next frame from equations (\ref{EQN:LIN}). Therefore,
we use more landmarks and more frames to determine the position of the next frame.
In theory, we can obtain the solution by using only two landmarks. The
corresponding formula is shown as the following equations:
\begin{equation}\label{EQN:2LD}
\left[
\begin{array}{cccccccc}
   0 & 0 & 1 & 0 & \frac{x_{p1}^{k}}{f} & 0 & 0 & 0 \\
   0 &0 &0 &1 & \frac{y_{p1}^{k}}{f} &0 &0 &0 \\
   1 &0 &-1 &0 & -\frac{x_{p1}^{k+1}}{f} &0 &0 &0\\
   0 &1 &0 &-1 & -\frac{y_{p1}^{k+1}}{f} &0 &0 &0\\
   0 &0 &0 &0 &0 &1 &0 & \frac{x_{p2}^{k}}{f}\\
   0 &0 &0 &0 &0 &0 &1 & \frac{y_{p2}^{k}}{f}\\
   1 &0 &0 &0 &0 &-1 &0 & -\frac{x_{p2}^{k+1}}{f}\\
   0 &1 &0 &0 &0 &0 &-1 & -\frac{y_{p2}^{k+1}}{f}\\
   \end{array}
\right]
\left[
\begin{array}{cccccccc}
  x_{k+1}\\
  y_{k+1}\\
  x_{l1}\\
  y_{l1}\\
  h_{1}^{k}\\
  x_{l2}\\
  y_{l2}\\
  h_{2}^{k}\\
\end{array}
\right]
  =
\left[
\begin{array}{ccccccc}
  x_{k}\\
  y_{k}\\
  \frac{\delta{h_{k}^{k+1}}}{f}x_{p1}^{k+1}\\
  \frac{\delta{h_{k}^{k+1}}}{f}y_{p1}^{k+1}\\
  x_{k+1}\\
  y_{k+1}\\
  \frac{\delta{h_{k}^{k+1}}}{f}x_{p2}^{k+1}\\
  \frac{\delta{h_{k}^{k+1}}}{f}y_{p2}^{k+1}\\
\end{array}
\right].
\end{equation}
From linear equations \eqref{EQN:2LD}, it is not difficult to obtain the position
of the next frame for the general case. For the convenience of subsequent
presentations, we represent the system of equations (\ref{EQN:2LD}) as
\begin{align}
    As = b \label{MTXEQN}.
\end{align}
\indent Note that the linear system of equations \eqref{EQN:2LD} is singular
when there is no height difference between two frames. In order to analyze
the singularity of the linear system of equations \eqref{EQN:2LD}, we simplify
it and obtain the following equivalent formula:
\begin{equation}\label{EQN2H}
 \begin{array}{lr}
    \left(x_{pn}^{k+1}-x_{pn}^{k}\right)\frac{h_{n}^{k}}{f} -
    \left(x_{p(n+1)}^{k+1}-x_{p(n+1)}^{k}\right)\frac{h_{n+1}^{k}}{f}
      = \left(x_{p(n+1)}^{k+1}-x_{pn}^{k+1}\right)\delta{h_{k}^{k+1}}, \\
    \left(y_{pn}^{k+1}-y_{pn}^{k}\right)\frac{h_{n}^{k}}{f} -
    \left(y_{p(n+1)}^{k+1}-y_{p(n+1)}^{k}\right)\frac{h_{n+1}^{k}}{f}
      = \left(y_{p(n+1)}^{k+1}-y_{pn}^{k+1}\right) \delta{h_{k}^{k+1}}.
\end{array}
\end{equation}
It is not difficult to see that it does not only determine landmark height
variables $h_{n}^{k}$ and $h_{n+1}^{k}$ from the linear system of
equations \eqref{EQN2H} when the height difference $\delta{h_{k}^{k+1}} = 0$.

\vskip 2mm

In order to overcome its singularity, we add the distance information between
two sequential frames to assist visual localization, which is provided by the
acceleromete in INS. Furthermore, we take into account the constant error and
the random walk error of the IMU, the angular line error caused by motion blur
and the error of the barometer. Then, the visual-inertial odometer problem is
modelled as a stochastic constrain optimization problem. The objective function
is formulated as follows:
\begin{align}
    \min \hskip 2mm \left( \left(\left(x_{k+1}-x_{k}\right)^2
    +\left(y_{k+1}-y_{k}\right)^2\right) -
     \left(\left(d_{k}^{k+1}\right)^2-\left(\delta{h_{k}^{k+1}}\right)^2\right)
     \right)^2 \nonumber \\
     +\left( \left(\left(x_{k+1}-x_{k-1}\right)^2
    +\left(y_{k+1}-y_{k-1}\right)^2\right) -
     \left(\left(d_{k-1}^{k+1}\right)^2-\left(\delta{h_{k-1}^{k+1}}\right)^2\right)
     \right)^2. \label{OPTOF}
\end{align}
For convenience, we denote the above objective function of equation
\eqref{OPTOF} as $f(s)$.

\vskip 2mm

We use the relationship about five landmarks and three frames, which has a
similar form to equation \eqref{MTXEQN}, as the constraint condition. Due
to the measurement error, the constraint condition can be denoted as
\begin{align}
   A_{\varepsilon}s = b_{\varepsilon}, \label{RANDCON}
\end{align}
where $A_{\varepsilon}$ includes the vision error and the altimeter error,
$b_{\varepsilon}$ is the constant vector with error term. Thus, the odometer
problem is  reformulated as a random equality-constrained
optimization problem. Note that we do not solve the following equivalent
nonlinear least-square problem:
\begin{align}
  \min \hskip 2mm f(s) + \|A_{\varepsilon}s - b_{\varepsilon} \|^2. \label{UNLSQ}
\end{align}
Replacing it, we construct a semi-implicit continuation method with a
trust-region technique to directly solve the linearly
equality-constrained optimization problem \eqref{OPTOF}-\eqref{RANDCON}
(see \cite{Goh2011, LLH2007, LKLT2009} and \cite{LL2010, Luo2012, LLW2013} for the
semi-implicit continuation method solving an unconstrained optimization problem
and the smallest eigenvalue or generalized eigenvalue problem, respectively).

\vskip 2mm

\subsection{Semi-implicit Continuation Method for the Optimization Subproblem}

\vskip 2mm

In this subsection, we firstly give a semi-implicit continuation method with a
trust-region technique for the linearly equality-constrained optimization
subproblem \eqref{OPTOF}-\eqref{RANDCON}. According to its first-order
Karush-Kuhn-Tucker condition
\begin{align}
  \nabla_{s} L(s, \lambda) &= \nabla f(s) + A^{T}_{\epsilon} \lambda = 0,
                                             \label{FOKKTG} \\
  A_{\epsilon}s &= b_{\epsilon},             \label{FOKKTC}
\end{align}
we construct a continuous differential-algebraic gradient flow with index 2
as follows:
\begin{align}
  \frac{ds}{dt} & = - \nabla L_{s}(s, \lambda)
  = -\left(\nabla f(s) + A^{T}_{\epsilon} \lambda \right),  \label{DAGF} \\
  A_{\epsilon}s  &= b_{\epsilon},                     \label{LACON}
\end{align}
where the Lagrangian function is written as
\begin{align}
   L(s, \lambda) = f(s) + \lambda^{T}(A_{\epsilon}s - b_{\epsilon}).
      \label{LAGFUN}
\end{align}

\vskip 2mm

In order to solve the continuous vector $\lambda(t)$ in \eqref{DAGF}
-\eqref{LACON}, via differentiating its algebraic constraint \eqref{LACON}
in variable $t$ and using its differential equation \eqref{DAGF}, we obtain
\begin{align}
  A_{\epsilon}\frac{ds}{dt} = - A_{\epsilon}
  \left(\nabla f(s) + A^{T}_{\epsilon} \lambda \right)
  = - A_{\epsilon} \nabla f(s) - A_{\epsilon}A^{T}_{\epsilon} \lambda = 0.
    \label{DIFALGC}
\end{align}
Assuming that matrix $A_{\epsilon}$ has a full row rank, from
\eqref{DIFALGC}, we know that $\lambda$ satisfies
\begin{align}
  \lambda = - \left(A_{\epsilon}A^{T}_{\epsilon} \right)^{-1} A_{\epsilon}
  \nabla f(s). \label{LAMBDA}
\end{align}
Replacing $\lambda$ in \eqref{DAGF} with equation \eqref{LAMBDA}, we obtain
the ordinary differential gradient flow as follows:
\begin{align}
  \frac{ds}{dt} = - \left( I - A_{\epsilon}^{T} \left(A_{\epsilon}^{T}\right)^{+}
  \right) \nabla f(s), \label{ODGF}
\end{align}
where $\left(A_{\epsilon}^{T}\right)^{+} = \left(A_{\epsilon}A^{T}_{\epsilon}
\right)^{-1} A_{\epsilon}$ is the generalized inverse of matrix
$A_{\epsilon}^{T}$. In other words, we also obtain the continuous projection
gradient flow in references \cite{BB1989, Tanabe1980} via another approach.

\vskip 2mm

We denote
\begin{align}
  P  = I - A_{\epsilon}^{T} \left(A_{\epsilon}^{T}\right)^{+}.  \label{PROMAT}
\end{align}
Then, it is not difficult to verify $P^{2} = P$, i.e., $P$ is a projection matrix
and $P$ is an orthogonal projector onto the null space $\mathcal{N}(A)$. Using
this property and from (\ref{ODGF})-(\ref{PROMAT}), we obtain
\begin{align}
 \frac{df(s)}{dt} & = - \nabla f(s)^{T} \frac{ds}{dt}
 = - (\nabla f(s))^{T} P \nabla f(s) = -
 (\nabla f(s))^{T} P^{2} \nabla f(s) \nonumber \\
 & = - (P \nabla f(s))^{T}(P \nabla f(s)) = - \|P \nabla f(s) \|_{2}^{2} \le 0,
 \nonumber
\end{align}
namely, the objective function $f(s)$ is decreasing along the
solution $s(t)$ of the continuous dynamical system \eqref{ODGF}. Furthermore,
Tanabe \cite{Tanabe1980} and Schropp \cite{Schropp2000} proves that the
solution $s(t)$ tends to $s^{*}$ as $t \to \infty$, where $s^{*}$ satisfies the
first-order Krarush-Kuhn-Tucker condition \eqref{FOKKTG}-\eqref{FOKKTC}. Thus,
We can expect to obtain an approximation solution of \eqref{OPTOF}-\eqref{RANDCON}
via following the trajectory of the ordinary differential dynamical system \eqref{ODGF} or
the differential-algebraic dynamical system \eqref{DAGF}-\eqref{LACON}.

\vskip 2mm

For the system of differential-algebraic equations \eqref{DAGF}-\eqref{LACON},
we look the algebraic equation \eqref{LACON} as a degenerate differential
equation. Then, applying the implicit Euler method to the total system,
we obtain ( see \cite{AP1998, HW1996, LF2000} for the implicit Euler method)
\begin{align}
   s_{k+1} &= s_{k} - \Delta t_{k} \left( \nabla f(s_{k+1})
   + A_{\epsilon}^{T}\lambda_{k+1} \right),      \label{IEDD} \\
   A_{\epsilon}s_{k+1} &= b_{\epsilon}.      \label{IEDA}
\end{align}
Replacing $\nabla f(s_{k+1})$ with its first-order approximation
$\nabla f(s_{k}) + \Delta t_{k} \nabla^{2} f(s_{k})$ and
$\lambda_{k+1}$ with $\lambda_{k}$ in \eqref{IEDD}, respectively, we obtain the
predicted variable $s_{k+1}^{P}$ of the $(k+1)^{th}$ iteration variable
$s_{k+1}$ as follows:
\begin{align}
  \left(\frac{1}{\Delta t_{k}}I + G_{k} \right)d_{k}
 & = - p_{g_{k}}, \label{PRDK} \\
 s_{k+1}^{P} & = s_{k} + d_{k}, \label{PRSK1}
\end{align}
where $G_{k} = \nabla^{2} f(s_{k})$ and $p_{g_{k}} =
\nabla f(s_{k})+ A_{\epsilon}^{T}\lambda_{k}$. Since the predicted point $s_{k+1}^{P}$
will escape from the constraint plane \eqref{LACON}, we pull it back via a
projection method as follows:
\begin{align}
   &\min \hskip 2mm  \left\|s - s_{k+1}^{P}\right\| \nonumber \\
   &\text{s.t.} \hskip 2mm  A_{\epsilon} s = b_{\epsilon}.  \label{MINDIST}
\end{align}
Using the Lagrangian multiplier method (see p. 479, \cite{Bertsekas2018}), it
is not difficult to obtain the solution of the projection problem \eqref{MINDIST}
as follows:
\begin{align}
 s_{k+1} = s_{k+1}^{P}
 + A_{\epsilon}^{T}\left(A_{\epsilon}A_{\epsilon}^{T}\right)^{-1}
 \left(b_{\epsilon} - A_{\epsilon}s_{k+1}^{P}\right).   \label{ITK1SK1}
\end{align}
Since $s_{k}$ is in the constraint plane \eqref{LACON}, from equations
\eqref{PRSK1} and \eqref{ITK1SK1}, we have
\begin{align}
  s_{k+1} & = s_{k} + d_{k}
  + A_{\epsilon}^{T}\left(A_{\epsilon}A_{\epsilon}^{T}\right)^{-1}
  \left(A_{\epsilon}s_{k} - A_{\epsilon}s_{k+1}^{P}\right) \nonumber \\
  & = s_{k} + d_{k}
  - A_{\epsilon}^{T}\left(A_{\epsilon}A_{\epsilon}^{T}\right)^{-1}
  A_{\epsilon}d_{k} \nonumber \\
  & = s_{k} + \left(I - A_{\epsilon}^{T}\left(A_{\epsilon}A_{\epsilon}^{T}\right)^{-1}
  A_{\epsilon} \right)d_{k} \nonumber \\
  & = s_{k} + Pd_{k}, \label{PRODK}
\end{align}
where projection matrix $P$ is defined by \eqref{PROMAT}.

\vskip 2mm

Using the implicit relationship \eqref{LAMBDA} between the Lagrangian multiplier
$\lambda(t)$ and the differential variable $s(t)$, we obtain the Lagrangian
multiplier at the $(k+1)^{th}$ iteration
\begin{align}
   \lambda_{k+1} = - \left(A_{\epsilon}A^{T}_{\epsilon} \right)^{-1} A_{\epsilon}
  \nabla f(s_{k+1}).  \label{LAMBDAK1}
\end{align}
Replacing equation \eqref{LAMBDAK1} into $p_{g_{k}} = \nabla f(s_{k})
+ A_{\epsilon}^{T}\lambda_{k}$, we have
\begin{align}
 p_{g_k} = \nabla f(s_{k})+ A_{\epsilon}^{T}\lambda_{k}
 = \left(I -A_{\epsilon}^{T}\left(A_{\epsilon}A^{T}_{\epsilon} \right)^{-1}
 A_{\epsilon}\right) \nabla f(s_{k}) = Pg_{k}, \label{PROGK}
\end{align}
where $g_{k} = \nabla f(s_{k})$.

\vskip 2mm

Another issue is how to adaptively adjust the time-stepping length $\Delta t_k$
every iteration. We borrow it from the trust-region method for its robust
global convergence property and fast local convergence property (see pp. 561-593,
\cite{SY2006}). Since variable $s_{k}$ is feasible, i.e. it always stay in the
constraint plane \eqref{LACON} every iteration, the objective function $f(s)$ is
a suitable merit function for adjusting the time stepping length $\Delta t_k$
as we use a trust-region technique.

\vskip 2mm

When a trust-region technique is selected, we need to construct an local
approximation model around variable $s_{k}$. According to the traditional approach,
we adopt a quadratic model as follows:
\begin{align}
  q_k(s) = (s-s_{k})^{T}\nabla f(s_k) + \frac{1}{2}(s-s_k)^{T}
      \nabla^{2} f(s_k) (s-s_{k}). \label{SOAM}
\end{align}
Now, based on the following measurement ratio
\begin{align}
 \rho_k = \frac{f(s_k)-f(s_{k+1})}{q_k(s_k)-q_k(s_{k+1})},
 \label{MRHOK}
\end{align}
we give an adaptive adjustment time-stepping length formula
\begin{equation}\label{ADTK1}
          \Delta t_{k+1} = \left\{
          \begin{array}{rl}
          \gamma_1 \Delta t_k, &{if \hskip 1mm 0 \leq \left|1- \rho_k \right| \le \eta_1,}\\
          \Delta t_k, &{if \hskip 1mm \eta_1 < \left|1 - \rho_k \right| < \eta_2,}\\
          \gamma_2 \Delta t_k, &{if \hskip 1mm \left|1-\rho_k \right| \geq \eta_2,}
          \end{array}\right.
\end{equation}
where the constants are selected as $0< \gamma_2 \le 1/2, \; 1< \gamma_1 \le 2$,
$0 < \eta_1 \le 0.25$ and $0.75 < \eta_2 < 1$ according to numerical experiments.
The specific algorithm steps are shown in algorithm \ref{alg:1}.

\vskip 2mm

\begin{algorithm}
	\renewcommand{\algorithmicrequire}{\textbf{Input:}}
	\renewcommand{\algorithmicensure}{\textbf{Output:}}
    \newcommand{\algorithmicbreak}{\textbf{break}}
    \newcommand{\BREAK}{\STATE \algorithmicbreak}
	\caption{Semi-implicit Continuation Method with
    Trust-region Technique for Linearly Equality-constrained Optimization}
	\label{alg:1}
	\begin{algorithmic}[1]
		\REQUIRE ~~\\
        An objective function: $f(s)$, \\
        and the linear constraint: $A_{\epsilon}s = b_{\epsilon}$, \\
        and the minimum absolute gradient bound of Lagrangian function
        $L(s,\lambda)= f(s)+\lambda^{T} (A_{\epsilon}s-b)$:  $\delta_\epsilon$.
		\ENSURE ~~\\
        The optimal approximation solution $s^{*}$.

        \vskip 2mm
        		
        \STATE Initialize a point $s_0$ and the parameter $\Delta t_0$.
        \STATE Choose constants $\eta_{a}, \eta_1,\eta_2,\gamma_1,\gamma_2$ to satisfy \\
        $0< \eta_{a} < \eta_1 \leq 1/2<\eta_2<1$  and  $0<\gamma_2 <1<\gamma_1$, \\
        such as $\eta_{a} = 10^{-6}, \; \eta_1 = 0.25, \; \eta_2 = 0.75$
        and $\gamma_1 = 2, \; \gamma_2 = 0.5$.
        \STATE $k \leftarrow 0$
        \STATE Compute $f_0 = f(s_0)$, $g_0 = \nabla f(s_0)$, $p_{g_{0}} =
        \nabla_{s} L(s_0,\lambda_0) = g_{0} + A^{T}_{\epsilon}\lambda_0$ and
        $G_0 = \nabla^{2}_{s} L(s_0, \lambda_0) = \nabla^{2} f(s_0)$, where the
        Lagrangian multiplier $\lambda_0 = -\left(A_{\epsilon}A^{T}_{\epsilon}\right)^{-1}
        A_{\epsilon}g_{0}$.
        \WHILE{$\|g_k\|> \delta_{\epsilon}$}
        \IF{$1/\Delta t_{k}I + G_k \succ 0$
        and $\left(1/\Delta t_{k}I + G_{k}- P^{T}G_{k}P \right) \succ 0$}
            \STATE Compute $d_{k}$ based on equation \eqref{PRDK}.
            \STATE Let $s_{k+1}^{P} = s_{k} + d_{k}$ and project $s_{k+1}^{P}$ to
            the constraint plane
            $A_{\epsilon}s = b_{\epsilon}$
            by solving problem \eqref{MINDIST}, and obtain
            $s_{k+1}$ which is given by equation \eqref{ITK1SK1}.
            \STATE Compute $f_{k+1} = f(s_{k+1})$ and the measurement ratio $\rho_{k}$ based on
            equations \eqref{SOAM}-\eqref{MRHOK}.
        \ELSE
        \STATE $\rho_k  = -1$.
        \ENDIF
        \IF{$\rho_k\le \eta_{a}$}
            \STATE $s_{k+1} = s_{k}$.
        \ELSE
            \STATE Accept $s_{k+1}$ and compute
            $g_{k+1} = \nabla f(s_{k+1})$, $G_{k+1} = \nabla^{2} f(s_{k+1})$,
            $\lambda_{k+1} = \left(A_{\epsilon}A^{T}_{\epsilon}\right)^{-1}A_{\epsilon}g_{k+1}$,
            and the projection gradient
            $p_{g_{k+1}} = g_{k+1} + A_{\epsilon}^{T}\lambda_{k+1}$.
        \ENDIF

        \STATE Adjust the time-stepping length $\Delta t_{k+1}$ based on the
        trust-region technique \eqref{ADTK1}.
        \STATE Update $\lambda_k \leftarrow \lambda_{k+1}$, $s_{k} \leftarrow s_{k+1}$,
        $f_{k} \leftarrow f_{k+1}$, $g_{k} \leftarrow g_{k+1}$, $G_{k} \leftarrow G_{k+1}$,
        $p_{g_{k}} \leftarrow p_{g_{k+1}}$ and $k \leftarrow k+1$.
        \ENDWHILE
	\end{algorithmic}
\end{algorithm}

\vskip 2mm

\subsection{Convergence Analysis of Semi-implicit Continuation Method for
optimization subproblem}

In this subsection, we give the local and the global convergence properties
of the semi-implicit continuation method for the linearly equality-constrained
optimization subproblem (i.e. Algorithm \ref{alg:1}). Firstly, we give an
estimation of upper bounds for the quadratic model $q_{k}(s_{k+1})$ which is
similar result of the trust-region method for unconstrained optimization
problem \cite{Powell1975}.

\vskip 2mm

\begin{lemma} \label{LBSOAM}
Assume that the quadratic model $q_{k}(s)$ is defined by \eqref{SOAM} and
$d_{k}$ is solved by equations \eqref{PRDK}-\eqref{ITK1SK1}. If
$\left(1/\Delta t_{k}I + G_{k} \right) \succ 0$ and
$\left(1/\Delta t_{k}I + G_{k}- P^{T}G_{k}P \right) \succ 0$ for some
$\Delta t_{k} > 0$, where projection matrix $P$ is given by equation
\eqref{PROMAT}, we have an estimation of lower bounds for the predicted
reduction $Pred_{k} = q_{k}(s_{k}) - q_{k}(s_{k}+Pd_{k})$ as follows:
\begin{align}
 Pred_{k} \ge \frac{1}{2} \left\|p_{g_{k}} \right\|
 \min\left\{\left\|Pd_{k}\right\|, \; \|p_{g_{k}}\|/\|G_{k}\|\right\},
 \label{PLBREDST}
\end{align}
where $p_{g_{k}} = \nabla_{s} L(s_{k}, \lambda_{k}) =
\nabla f(s_{k}) + A_{\epsilon}^{T}\lambda_{k}$ and the Lagrange multiplier
$\lambda_{k}$ is determined by equation \eqref{LAMBDAK1}.
\end{lemma}
\proof Assume that $d_{k}$ is the solution of equation \eqref{PRDK}.
Then, we have
\begin{align}
  q_{k}(s_{k}) &- q_{k}(s_{k}+Pd_{k}) = -\frac{1}{2}d_{k}^{T}P^{T}G_{k}Pd_{k}
  - g_{k}^{T}Pd_{k}  \nonumber \\
  & = -\frac{1}{2}d_{k}^{T}P^{T}G_{k}Pd_{k}
  + p_{g_k}^{T} \left(\mu_{k}I + G_{k}\right)^{-1}p_{g_k}  \nonumber \\
  &=  \frac{1}{2}p_{g_k}^{T} \left(\mu_{k}I + G_{k}\right)^{-1}p_{g_k}
  + \frac{1}{2}d_{k}^{T}\left(-P^{T}G_{k}P + \mu_{k}I + G_{k}\right)d_{k},
  \label{DKGKDK}
\end{align}
where we denote $\mu_{k} = 1/\Delta t_{k}$. From the above equality
\eqref{DKGKDK}, $\left(\mu_{k} I + G_{k} \right) \succ 0$ and selecting a
constant $\mu_{lb}$ such that $\mu_{lb} = \min\left\{0, -\lambda_{min}
\left(G_{k}-P^{T}G_{k}P\right)\right\}$, where $\lambda_{min}
\left(G_{k}-P^{T}G_{k}P\right)$ is the smallest eigenvalue of matrix
$\left(G_{k}-P^{T}G_{k}P\right)$, we obtain
\begin{align}
 q_{k}(s_{k}) &- q_{k}(s_{k+1}) \ge \frac{1}{2}p_{g_k}^{T}
 \left(\mu_{k}I + G_{k}\right)^{-1}p_{g_k} + \frac{1}{2}
 \left(\mu_{k} - \mu_{lb}\right)\left\|d_{k}\right\|^{2} \nonumber \\
 & \ge \frac{1}{2}\left(\frac{1}{\mu_{k}+\left\|G_{k}\right\|}
 {\left\|p_{g_{k}}\right\|^{2}}
 + \left(\mu_{k} - \mu_{lb}\right)\left\|d_{k}\right\|^{2} \right). \label{LBEQK}
\end{align}

\vskip 2mm

Now we consider the properties of the function
\begin{align}
 \varphi(\mu) \equiv \mu \left\|d_k\right\|^{2}
 + \frac{1}{\mu + \mu_{lb}+\left\|G_k \right\|} \left\|p_{g_k}\right\|^2.
 \label{VLF}
\end{align}
It is not difficult to know that the function $\varphi(\mu)$ is
convex when $\left(\mu + \|G_{k}\right\|) > 0$, since $\varphi^{''}(\mu)
= 2\|p_{g_k}\|^2 /\left(\mu + \mu_{lb} + \|G_{k}\|\right)^3  \ge 0$. Thus,
the function $\varphi(\mu)$ attains its minimizer $\varphi(\mu_{min})$ when
$\mu_{min}$ satisfies $\varphi^{'}(\mu_{min})=0$ and
$\mu \ge -(\mu_{lb}+\|G_k\|)$, i.e.
\begin{align}
  \varphi(\mu_{min}) = 2\|p_{g_k}\| \|d_k\|
  - \left(\mu_{lb}+\|G_k\|\right) \|d_k\|^{2},   \label{MINV}
\end{align}
when
\begin{align}
 \mu_{min} = \|p_{g_k}\|/\|d_k\| - \mu_{lb}-\|G_k\|, \; \text{and} \;
 \mu_{min} > -\left(\mu_{lb}+\|G_k\|\right). \label{MINLD}
\end{align}

\vskip 2mm

We prove the property \eqref{PLBREDST} by distinguishing two cases
separately, namely $\mu_{min}$ is nonnegative or negative. When $\|p_{g_k}
\|/\|d_k\| \ge \left(\mu_{lb}+\|G_k\|\right)$, from (\ref{MINLD}), we have
$\mu_{min} \ge 0$. For this case, combining $\mu_k \ge \mu_{lb}$ with equations
\eqref{LBEQK}--\eqref{MINLD}, we obtain
\begin{align}
 q_k(s_{k}) & - q_k(s_{k}+Pd_{k})  \ge (\mu_{k}-\mu_{lb}) \|d_k\|^{2} +
 \frac{1}{\mu_{k} + \|G_k\|} \|g_k\|^2
 =  \varphi(\mu_{k}-\mu_{lb})
 \ge \varphi(\mu_{min}) \nonumber \\
 & = \frac{1}{2} \left(\|p_{g_k}\| \|d_k\| + \left(\|p_{g_k}\| \|d_k\|-
 \left(\mu_{lb}+\|G_k\|\right)\|d_k\|^{2}\right)\right)
 \ge \frac{1}{2} \|p_{g_k}\| \|d_k\|.   \label{MINQCOV}
\end{align}

\vskip 2mm

The other case is ${\|p_{g_k}\|}/{\|d_k\|} < \left(\mu_{lb}+\|G_k\|\right)$,
which gives $\mu_{min} < 0$ from (\ref{MINLD}). Since the function
$\varphi(\mu)$ is monotonically increasing for all $\mu \ge 0$ when
$\|p_{g_k}\|/\|d_k\| < \left(\mu_{lb}+\|G_k\|\right)$, from equations
\eqref{LBEQK}--\eqref{VLF}, we obtain
\begin{align}
 q_k(s_{k}) & - q_k(s_{k}+Pd_{k})
  \ge \frac{1}{2}\left((\mu_{k}-\mu_{0}) \|d_k\|^{2} +\frac{1}
  {\mu_k + \|G_k \|} \|p_{g_k}\|^2\right) \nonumber \\
  &= \frac{1}{2} \varphi(\mu_{k}-\mu_{lb}) \ge \frac{1}{2} \varphi(0)
  = \frac{1}{2} \|p_{g_k}\|^2/\|G_k\|.  \label{MINQMI}
\end{align}

\vskip 2mm

Combining (\ref{MINQCOV}) and (\ref{MINQMI}), we get
\begin{align}
   q_k(s_{k}) - q_k(s_{k}+Pd_{k}) \ge \frac{1}{2} \|p_{g_k}\|
   \min\left\{\|p_{g_k} \|/\|G_k\|, \; \|d_k\|\right\}. \label{MINQATR}
\end{align}
Since $s_{k+1}$ is the projection of $s_{k+1}^{P} = s_{k} + d_{k}$ in a convex
set $\mathrm{C_{s}} = \{s: A_{\epsilon}s = b_{\epsilon}\}$,  according to Projection Theorem
(see Proposition 1.1.4, p. 19 \cite{Bertsekas2018}), we have
\begin{align}
 \|Pd_{k}\| = \|Ps_{k+1}^{P} - Ps_{k}\| \le \|s_{k+1}^{P} - s_{k}\|
 = \|d_{k}\|. \label{PROTHE}
\end{align}
Using inequality \eqref{PROTHE} in equation \eqref{MINQATR}, we obtain an
estimation \eqref{PLBREDST}, which proves the lemma. \eproof

\vskip 2mm

In order to prove that $p_{g_k}$ tends to zero, we also use the following result
about the lower bound estimation of the time-stepping length $\Delta t_{k}$ when
$\|p_{g_k}\| \ge \delta_{p_g} > 0$.

\vskip 2mm

\begin{lemma} \label{DTBOUND}
Assume that the level set of the twice continuously differentiable function
$f: \; \mathrm{R}^{n} \to \mathrm{R}$ in the linear constraint plane
\eqref{LACON} is bounded, i.e. $L_{f} = \{s: \; f(s) \le f(s_0), \;
A_{\epsilon}s = b_{\epsilon}\}$ is bounded. Furthermore, assume that there
exists a positive constant $\delta_{g}$ such that
\begin{align}
 \|p_{g_{k}}\| \ge \delta_{p_g} > 0, \; k = 1, \; 2, \dots \label{PGKGEPN}
\end{align}
are satisfied, where $p_{g_{k}}$ are generated by Algorithm \ref{alg:1}.
Then, it exists a positive $\delta_{\Delta t}$ such that the time-stepping
length
\begin{align}
  \Delta t_{k} \ge \delta_{\Delta t} > 0, \; k = 1, \; 2, \dots \label{DTGEPN}
\end{align}
are satisfied, where $\Delta t_{k}$ is adaptively adjusted by formula
\eqref{ADTK1}.
\end{lemma}

\vskip 2mm

\proof Since the level set $L_{f}$ is bounded, according to Proposition A.7 in
pp. 754-755 of reference \cite{Bertsekas2018}, $L_{f}$ is closed. Then, there
exists two positive constants $M_{p_{g}}$ and $M_{G}$ such that
\begin{align}
  \|p_{g_{k}}\| \le M_{p_{g}}, \; \|G_{k}\| \le M_{G}, \; k = 1, \; 2, \dots
  \label{PGGKBD}
\end{align}
are satisfied, respectively. Selecting a positive $\delta_{\Delta t_{0}}
= 1/(2M_{G})$, we have $\left(1/\Delta t_{k}I + G_{k} \right) \succ 0$ and
$\left(1/\Delta t_{k}I + G_{k}- P^{T}G_{k}P \right) \succ 0$
when $\Delta t_{k} \le \delta_{\Delta t_{0}}$, where projection
matrix $P$ is given by equation \eqref{PROMAT}.

\vskip 2mm

From equations \eqref{MRHOK}, \eqref{PGGKBD} and the reduction estimation
\eqref{PLBREDST} of the quadratic model (see Lemma \ref{DTBOUND}),
when $\Delta t_{k} \le \delta_{\Delta t_{0}}$, we obtain the estimation of the
measurement ratio
\begin{align}
 \left|\rho_{k} - 1\right| &=  \left|\frac{(f(s_{k}) - f(s_{k}+Pd_{k}))
 - (q_{k}(s_{k}) - q_{k}(s_{k}+Pd_{k}))}{q_{k}(s_{k}) - q_{k}(s_{k}+Pd_{k})}\right|
 \nonumber \\
 &= \left|\frac{1/2(Pd_{k})^{T}\left(G_{k} - \nabla^2 f(\bar{s_{k}})\right)(Pd_{k})}
 {q_{k}(s_{k}) - q_{k}(s_{k}+Pd_{k})}\right|
 \le \frac{M_{G}\left\|Pd_{k}\right\|^2}{|q_{k}(s_{k}) - q_{k}(s_{k}+Pd_{k})|}
 \nonumber \\
 & \le \frac{M_{G}\left\|Pd_{k}\right\|^2}
 {\frac{1}{2} \left\|p_{g_{k}} \right\|
 \min\left\{\left\|Pd_{k}\right\|, \; \|p_{g_{k}}\|/\|G_{k}\|\right\}}
 \le \frac{2M_{G}\left\|Pd_{k}\right\|^2}
 {\delta_{p_g} \min\left\{\left\|Pd_{k}\right\|, \; \delta_{p_g}/M_{G}\right\}}.
 \label{ESTRHOK}
\end{align}
In the above third inequality and the last inequality, we use the Cauchy-Schwartz
inequality $|x^{T}y|\le \|x\|\|y\|$ and the lower bound assumption
\eqref{PGKGEPN} of the projection gradient $p_{g_{k}}$.

\vskip 2mm

Selecting a positive constant $\delta_{\Delta_{t_{1}}}
= \min\{\delta_{p_{g}}/M_{G}, \; \eta_{1}\delta_{p_{g}}/(2M_{G})\}$, when
$\Delta t_{k} \le \delta_{\Delta t_{0}}$ and
$\left\|Pd_{k}\right\| \le \delta_{\Delta_{t_{1}}}$, from equation
\eqref{ESTRHOK}, we have
\begin{align}
  \left|\rho_{k} - 1\right| \le \eta_{1}, \label{RHOLETA1}
\end{align}
which means that the predicted point $s_{k+1} = s_{k} + Pd_{k}$ is accepted
and the time-stepping length $\Delta t_{k+1}$ is enlarged according to the
time-stepping adjustment formula \eqref{ADTK1}.

\vskip 2mm

From equations \eqref{PRDK} and \eqref{PGGKBD},
when $\Delta t_{k} \le \delta_{\Delta t_2} =
\min\{1/\delta_{\Delta t_0}, \; 1/(M_{p_g}/\delta_{\Delta t_1} + M_{G})\}$,
we have
\begin{align}
 \left\|d_{k}\right\| &= \left\|\left(\frac{1}{\Delta t_{k}}I
 + G_{k} \right)^{-1}p_{g_{k}}\right\| \nonumber \\
 &\le \frac{\|p_{g_{k}}\|}{1/\Delta t_{k} - \|G_{k}\|}
 \le \frac{M_{p_{g}}}{1/\Delta t_{k} - M_{G}} \le \delta_{\Delta t_{1}},
 \label{DKLECON}
\end{align}
which means that inequality \eqref{RHOLETA1} is satisfied according to the
projection property \eqref{PROTHE} (i.e., $\|Pd_{k}\| \le \|d_{k}\|$).

\vskip 2mm

Assume that $K$ is the first index such that $\Delta t_{K} \le
\delta_{\Delta t_2}$ is satisfied. Then, according to the projection property
\eqref{PROTHE}, inequalities \eqref{DKLECON} and \eqref{RHOLETA1}, we know
that $|\rho_{K} - 1 | \le \eta_{1}$, which means that $s_{K} + Pd_{K}$ is
accepted and the time-stepping length $\Delta t_{K+1}$ is enlarged according
to the time-stepping adjustment formula \eqref{ADTK1}. Consequently, the
time-stepping length $\Delta t_{k}\ge \min \{\gamma_{1} \delta_{\Delta t_2},
\; \Delta t_{K}\}$ when $k \ge K$, which proves the lemma. \eproof

\vskip 2mm

Using the results of Lemma \ref{LBSOAM} and Lemma \ref{DTBOUND}, we can prove
the global convergence property of Algorithm \ref{alg:1} for a linearly
equality-constrained optimization subproblem.

\vskip 2mm

\begin{theorem}
Assume that the level set of the twice continuously differentiable function
$f(s)$ in the linear constraint plane \eqref{LACON} is bounded, i.e.
$L_{f} = \{s: f(s) \le f(s_0), \; A_{\epsilon}s = b_{\epsilon}\}$ is bounded.
Then, $\lim_{k \to \infty} \inf \|p_{g_{k}}\| = 0$, where
$p_{g_{k}} = \nabla f(s_{k}) + A_{\epsilon}^{T}\lambda_{k}$ and $s_{k}, \;
\lambda_{k}$ are generated by Algorithm \ref{alg:1}.
\end{theorem}
\proof We will prove it by contradiction. Assume that the conclusion is not true.
Then it exists a positive constant $\delta_{p_{g}}$ such that
\begin{align}
  \|p_{g_{k}}\| \ge \delta_{p_{g}} > 0, \;  k = 1, \; 2, \dots \label{PGKGECON}
\end{align}
are satisfied. According to Algorithm \ref{alg:1}, we know that it exists
an infinite subsequent $k_{i}$ such that trial step $Pd_{k_i}$
are accepted, i.e., $\rho_{k_{i}} \ge \eta_{a}$, which gives
\begin{align}
  f_{0} - \lim_{k \to \infty} f_{k} = \sum_{k = 0}^{\infty} (f_{k} - f_{k+1})
  \ge \eta_{a} \sum_{k_{i} = 0}^{\infty}
  \left(q_{k}(s_{k_i}) - q_{k}(s_{k_i}+Pd_{k_i})\right),   \label{LIMSUMFK}
\end{align}
where $Pd_{k}$ is computed by equations \eqref{PRDK} and \eqref{PRODK}.
Using the bounded assumption of the objective function $f(s)$ in
the level set $L_{f}$ for inequality \eqref{LIMSUMFK}, we know
\begin{align}
  \lim_{k \to \infty} \left(q_{k}(s_{k_i}) - q_{k}(s_{k_i}+Pd_{k_i})\right) = 0.
  \label{LIMQK}
\end{align}
From the result of Lemma \ref{LBSOAM}, i.e. inequality \eqref{PLBREDST},
and equation \eqref{LIMQK}, we get
\begin{align}
   \lim_{k_i \to \infty} \|p_{g_{k_i}}\|
   \min \left\{\|p_{g_{k_i}} \|/\|G_{k_i}\|, \; \|Pd_{k_i}\|\right\} = 0,
   \label{LIMPK}
\end{align}
where $G_{k_i} = \nabla^2 f(s_{k_i})$.

\vskip 2mm

According to the bounded assumption of the level set $L_{f}$,
there
exists two positive constants $M_{p_{g}}$ and $M_{G}$ such that
\begin{align}
  \|p_{g_{k}}\| \le M_{p_{g}}, \; \|G_{k}\| \le M_{G}, \; k = 1, \; 2, \dots
  \label{PGGKBD}
\end{align}
are satisfied, respectively. From equation \eqref{PGGKBD} and inequalities
\eqref{PGKGECON} and \eqref{PGGKBD}, for the subsequent $\{k_{i}\}$ of accepted
trial steps, we obtain
\begin{align}
  \lim_{k_i \to \infty} \|Pd_{k_i}\| = 0.   \label{LIMPDK}
\end{align}
According to the bounded assumption of $p_{g_k}$ \eqref{PGKGECON} and
inequality \eqref{PGGKBD}, from the result of Lemma \ref{DTBOUND}, we know
that it exists a positive constant $\delta_{\Delta t}$ such that
\begin{align}
 \Delta t_{k} \ge \delta_{\Delta t} > 0, \;  k = 1,\; 2, \dots
 \label{DELTKGE}
\end{align}
are satisfied.

\vskip 2mm

From equation \eqref{PRDK} and using the property
$P^{2} = P$ of projection matrix $P$ which is defined by \eqref{PROMAT}, we
obtain
\begin{align}
  Pd_{k_i} = \left(P \left(\frac{1}{\Delta t_{k_i}}I
  + G_{k_i}\right)^{-1}P\right)p_{g_{k_i}}, \label{PDKEQ}
\end{align}
which gives
\begin{align}
 p_{g_{k_i}}^{T}Pd_{k_i} & = p_{g_{k_i}}^{T}\left(P \left(\frac{1}{\Delta t_{k_i}}I
  + G_{k_i}\right)^{-1}P\right)p_{g_{k_i}} \nonumber \\
  & = p_{g_{k_i}}^{T}\left(\frac{1}{\Delta t_{k_i}}I
  + G_{k_i}\right)^{-1}p_{g_{k_i}} \ge \left\|p_{g_{k_i}}\right\|^2
  \frac{1}{1/\delta_{\Delta t} - M_{G}}. \label{PGKDK}
\end{align}
Using the Cauchy-Schwartz inequality $|x^{T}y| \le \|x\|\|y\|$, from equation
\eqref{PGKGECON}, we have
\begin{align}
 \frac{1}{1/\delta_{\Delta t} - M_{G}} \left\|p_{g_{k_i}}\right\|^2 \le
 |p_{g_{k_i}}^{T}Pd_{k_i}| \le \left\|p_{g_{k_i}}\right\| \left\|Pd_{k_i}\right\|
\end{align}
which gives
\begin{align}
   \left\|p_{g_{k_i}}\right\| \le \left(1/\delta_{\Delta t} - M_{G}\right)
    \left\|Pd_{k_i}\right\|. \label{PGKLEPDK}
\end{align}
From inequality \eqref{PGKLEPDK} and equation \eqref{LIMPDK}, we obtain
\begin{align}
  \lim_{{k_i} \to \infty}\left\|p_{g_{k_i}}\right\| = 0, \label{PGKTOZ}
\end{align}
which contradicts the lower bound assumption \eqref{PGKGECON}. Therefore,
we prove the  conclusion of the theorem.  \eproof

\vskip 2mm

\subsection{Visual-Inertial Algorithm Descriptions} \label{SubSecAlg}

\vskip 2mm

The proposed visual-inertial odometer method is described in
Algorithm \ref{alg:2}. To convention, the input INS reading has been pre-calibrated
and the camera intrinsic parameters have been obtained. In algorithm \ref{alg:2},
the positions of the $(k-1)^{th}$ and $k^{th}$ frame have been determined before,
denoted as $Pc_{k-1}$ and $Pc_{k}$. And let $dist_{k-1}^{k+1}$ and $dist_{k}^{k+1}$
be the distances between the previous two frames and the $(k+1)^{th}$ frame,
respectively, which are measured by INS. Similarly, let $\delta{h_{k-1}^{k+1}}$
and $\delta{h_{k}^{k+1}}$ present the altitude differences between the previous
two frames and the $(k+1)^{th}$ frame, respectively, which are obtained by
altimeter. Then, we use feature matching to obtain the landmarks' locations in
frame coordinate system. Finally, the $(k+1)^{th}$ frame position is determined
through solving the linearly equality-constrained optimization
problem \eqref{OPTOF}-\eqref{RANDCON}, which is solved by Algorithm
\ref{alg:1}. After a number of iterations, the aircraft trajectory is determined.

\vskip 2mm

We just consider the relationships between frames and and landmarks,
and the visual-inertial method have not loop closure detection. Therefore,
algorithm \ref{alg:1} has good real-time performance. Additionlly, the proposed
algorithm tolerates a certain level of altitude error as we take into account
the random error of each component.

\vskip 2mm

\begin{algorithm}
	\renewcommand{\algorithmicrequire}{\textbf{Input:}}
	\renewcommand{\algorithmicensure}{\textbf{Output:}}
	\caption{Visual-inertial Odometry Algorithm}
	\label{alg:2}
	\begin{algorithmic}[1]
		\REQUIRE ~~\\
         the $(k-1)^th$ and $k^th$ frames' locations $Pc_{k-1}$, $Pc_{k}$, respectively;
         the distances $dist_{k-1}^{k+1}$, $dist_{k}^{k+1}$ between the previous
         two frames and the $(k+1)^{th}$ frame, respectively; and the altitude
         differences $\delta{h_{k-1}^{k+1}}$, $\delta{h_{k}^{k+1}}$ between the previous
         two frames and the $(k+1)^{th}$ frame, respectively.
		\ENSURE ~~\\
        The next frame location $Pc_{k+1}$.

        \vskip 2mm

        \FOR{a number of iterations}
            \STATE Determining the landmarks $\leftarrow$ matching the ORB
            feature in the $(k+1)^{th}$ frame and the ORB features in the
            previous two frames, respectively.
        \ENDFOR
        \FOR{a number of iterations}
        \STATE Obtain the landmarks' locations in the $(k+1)^{th}$ frame coordinate
        system and in previous frame coordinate system.
        \ENDFOR
        \STATE Get matrix $A_{\epsilon}$ in equation (\ref{RANDCON}).
        \STATE Solve the equality-constrained optimization problem
        \eqref{OPTOF}-\eqref{RANDCON} with Algorithm \ref{alg:1}.
        \STATE \textbf{return}  $Pc_{k+1}$.
	\end{algorithmic}
\end{algorithm}

\vskip 2mm

\section{Simulation Results}

\vskip 2mm

In order to illustrate the effect of the proposed algorithm, we compare the
localization accuracy of our algorithm and the pure inertial navigation on the
same trajectory. The aircraft sets off at some point in the equator and then
flies along the equatorial plane for an hour. According to the given condition
from the industry, we assume that the aircraft flies an hour at an altitude of
1200 meters with speed 235 meters per second. We also consider the line-of-sight
angular error to be less than 0.2 degrees, the random error of altimeter and
the altimeter error associated with the distance. The specific
key parameters are shown in table \ref{tbl:DP}.

\vskip 2mm

\begin{table}[http]
  \centering
  \caption{Aircraft key performance parameters}\label{tbl:DP}
  \begin{tabular}{|c|c|}
  \hline
  \hline
  Description                                      & Parameter value                       \\
  \hline
  Flight altitude of the aircraft                  & $1000 \sim 1500$ meters               \\
  Flight speed of the aircraft                     & $210 \sim 260$ meters per second      \\
  The line-of-sight angular error of landmarks           & $\leq0.2^{\circ}$               \\
  The random error of altimeter                    & one meter (variance $\sigma$ value)   \\
  The altimeter error related to flight distance   & $<$ Flight distance $*0.0001$         \\
  Horizontal attitude error of INS                 & $< 0.06^{\circ}$                      \\
  The heading error of INS                         & $<0.4^{\circ}$                        \\
  Required accuracy of localization                & $<900$ meters per hour                \\
  \hline
  \end{tabular}
\end{table}

\vskip 2mm

\begin{figure}[!htbp]
\centering
  \begin{minipage}[b]{0.5\linewidth}
    %\centering
    \includegraphics[width=2.3 in,height =2.3in]{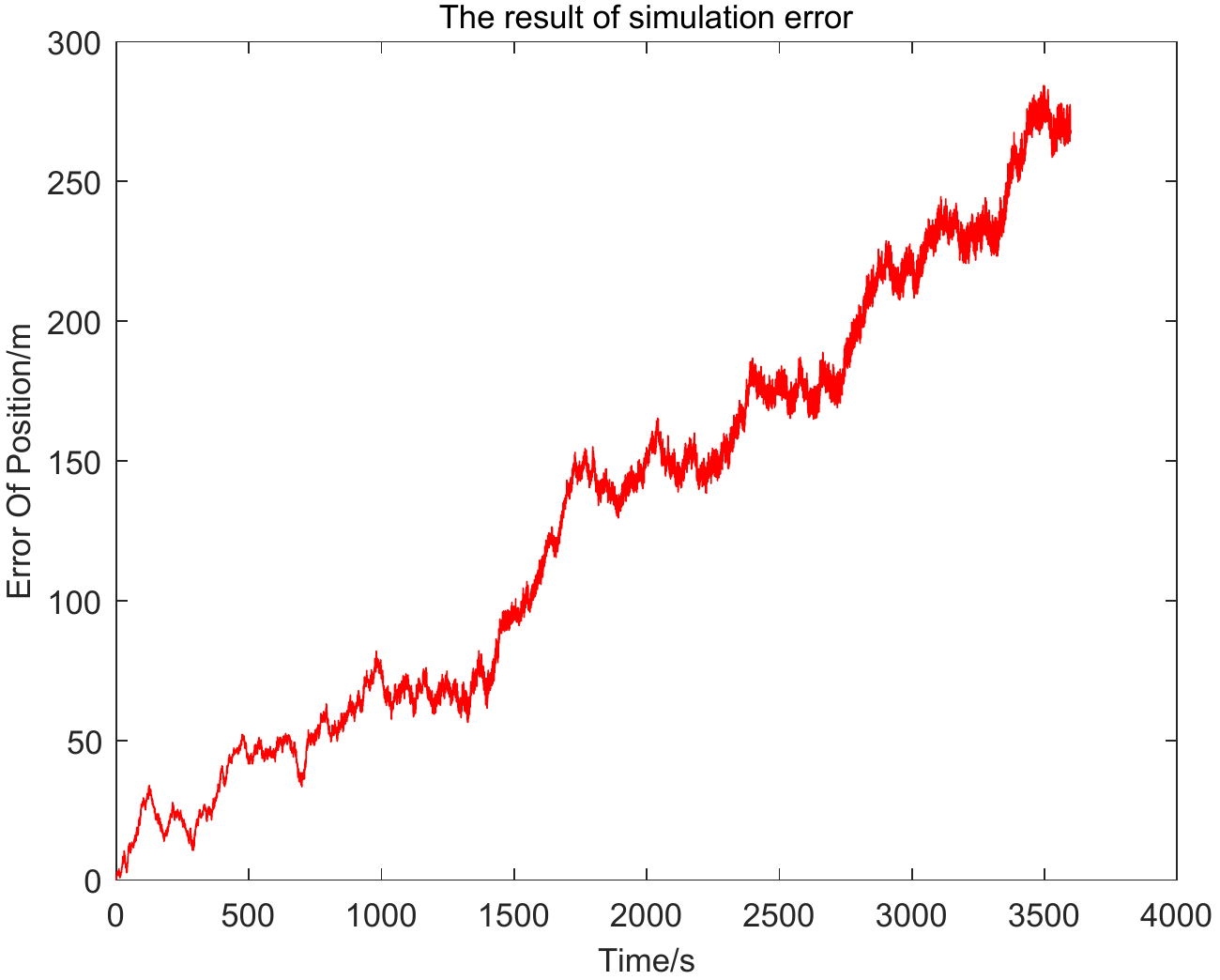}
    \caption{The result of proposed simulation positioning errors.}
    \label{Fig:RPSPE}
  \end{minipage}%
  \begin{minipage}[b]{0.5\linewidth}
    %\centering
    \includegraphics[width=2.3 in,height =2.3in]{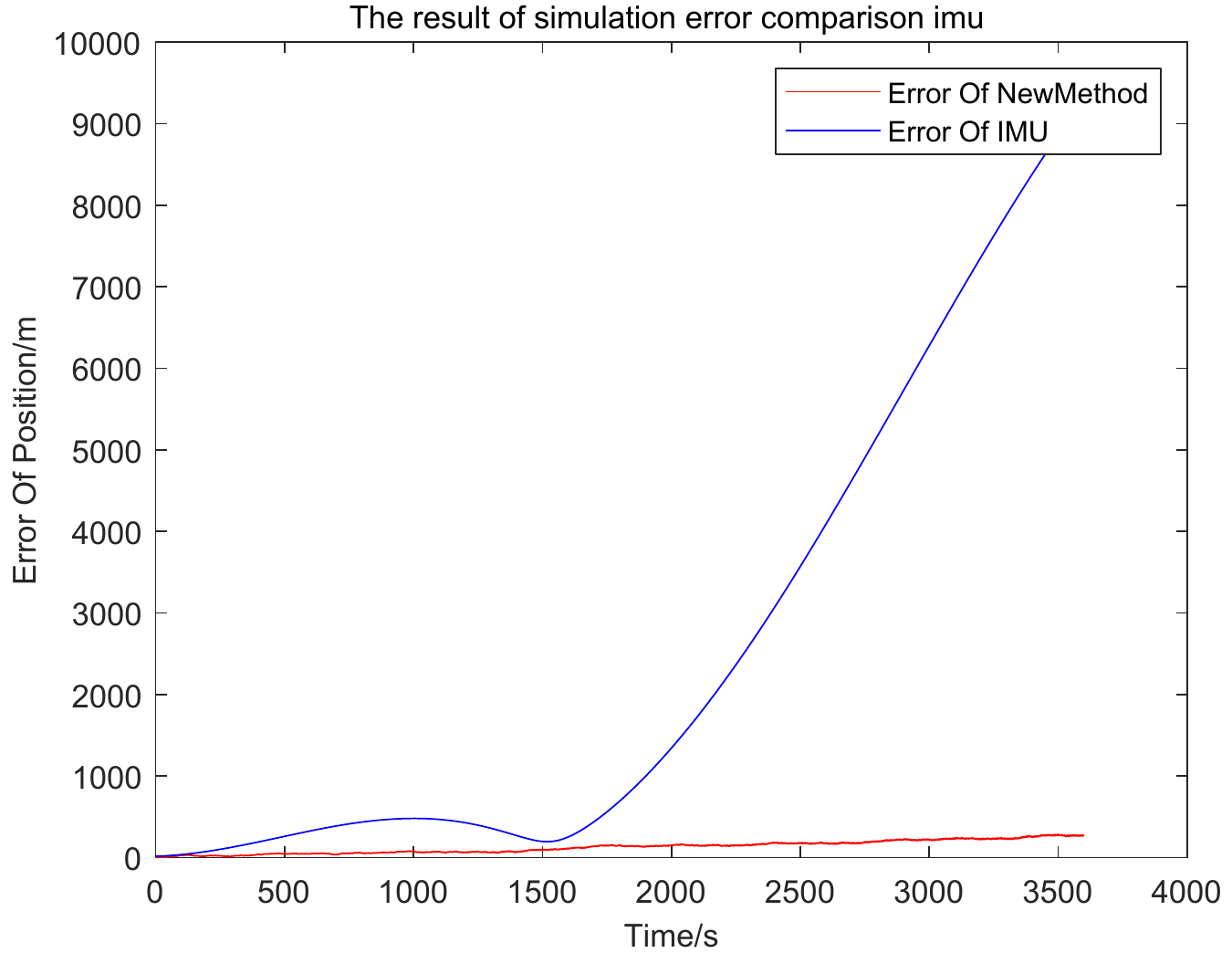}
    \caption{The comparison between proposed method and pure INS.}
    \label{Fig:CBPMPI}
  \end{minipage}
\end{figure}

The simulation error of the proposed method is shown in figure
\ref{Fig:RPSPE}. Figure \ref{Fig:CBPMPI} presents the comparison between our
method and the pure inertial navigation method. The vertical axis represents
the error between the real positioning location and the ideal trajectory. The
horizontal axis is about the flying time. Owing to the long-time high-speed
flight, the pure INS method does not work well.
From Figure \ref{Fig:CBPMPI}, we find that the error of inertial
navigation is more than 9 kilometers per hour, and our method which combines the
advantages of inertial navigation and visual odometry effectively suppress the
rapid propagation of errors. The accuracy of proposed method is only slightly less
than 300 meters. The effect of our proposed method has a noticeable
improvement and its accuracy meets the navigation accuracy requirements.

\vskip 2mm

\vskip 2mm

\section{Conclusion and Future Work}
The proposed algorithm combining the visual odometer to assist INS effectively
utilizes the complementarity of two methods. It avoids the rapid accumulation of
errors in an inertial navigation method, and has no problem of scale ambiguity.
Since there is no the loop closure detection, the proposed algorithm has good
real-time performance compared to other vision-based methods. Currently, we only
consider the horizonal flight with small variation in yaw angle. When the roll
angle and the pitch angle frequently change, the proposed method do not work very
well. Thus, the proposed algorithm is only applicable to the four DOF motion.
In order to solve the localization in this six DOF flight scenario,
we will design a more robust algorithm for full freedom navigation
base on the newly proposed method in the future.

\vskip 2mm

\vskip 2mm

\section*{Financial and Ethical Disclosures}
\begin{itemize}
  \item[$\bullet$]  Funding: This study was funded by by Grant 61876199 from National
  Natural Science Foundation of China, Grant YBWL2011085 from Huawei Technologies
  Co., Ltd., and Grant YJCB2011003HI from the Innovation Research Program of Huawei
  Technologies Co., Ltd..
  \item[$\bullet$]  Conflict of Interest: The authors declare that they have no
  conflict of interest.
\end{itemize}

\noindent \textbf{Acknowledgments} 
The first author is grateful to Professor Ya-Xiang Yuan and this work is
dedicated to him on the occasion of his 60th birthday.

\end{document}